\documentclass[letterpaper]{article} 

\usepackage{aaai2026}  
\nocopyright

\usepackage{times}  
\usepackage{helvet}  
\usepackage{courier}  
\usepackage[hyphens]{url}  
\usepackage{graphicx} 
\usepackage{natbib}  
\usepackage{caption} 
\usepackage{algorithm}
\usepackage{algpseudocode}
\usepackage{amsmath}
\usepackage{amsfonts}
\usepackage{amssymb}
\usepackage{ragged2e}
\usepackage{enumitem}
\usepackage{cleveref}
\usepackage{booktabs}
\usepackage{adjustbox}
\usepackage{tikz}
\usepackage{fontawesome5}
\usetikzlibrary{
    arrows.meta,
    positioning,
    shapes.multipart,
    shapes.geometric,
    shadows,
    calc,
    decorations.pathreplacing,
    decorations.pathmorphing,
    backgrounds
}
\usepackage{subcaption}
\usepackage{xcolor}
\definecolor{lightgray}{rgb}{0.96,0.96,0.96}
\usepackage{newfloat}
\usepackage{listings}
\usepackage{multirow}
\DeclareCaptionStyle{ruled}{labelfont=normalfont,labelsep=colon,strut=off} 
\floatstyle{ruled}
\newfloat{listing}{tb}{lst}{}
\floatname{listing}{Listing}
\title{Scalable Multi-Task Reinforcement Learning for Generalizable Spatial Intelligence in Visuomotor Agents}
\author{
    Shaofei Cai\equalcontrib\textsuperscript{\rm 1}, Zhancun Mu\equalcontrib\textsuperscript{\rm 1}, Haiwen Xia\textsuperscript{\rm 1}, Bowei Zhang\textsuperscript{\rm 1}, Anji Liu\textsuperscript{\rm 2}, Yitao Liang\textsuperscript{\rm 1}
}
\affiliations{
    \textsuperscript{\rm 1}Institute for Artificial Intelligence, Peking University\\
    \textsuperscript{\rm 2}School of Computing, National University of Singapore\\
    \{caishaofei, muzhancun, 2300010813, zhangbowei\}@stu.pku.edu.cn, anjiliu@nus.edu.sg, yitaol@pku.edu.cn
}

\usepackage{bibentry}

\begin{document}

\maketitle
\thispagestyle{plain} 
\pagestyle{plain} 

\begin{abstract}
While Reinforcement Learning (RL) has achieved remarkable success in language modeling, its triumph hasn't yet fully translated to visuomotor agents. A primary challenge in RL models is their tendency to overfit specific tasks or environments, thereby hindering the acquisition of generalizable behaviors across diverse settings. This paper provides a preliminary answer to this challenge by demonstrating that RL-finetuned visuomotor agents in Minecraft can achieve zero-shot generalization to unseen worlds. 
Specifically, we explore RL's potential to enhance generalizable spatial reasoning and interaction capabilities in 3D worlds. To address challenges in multi-task RL representation, we analyze and establish cross-view goal specification as a unified multi-task goal space for visuomotor policies. Furthermore, to overcome the significant bottleneck of manual task design, we propose automated task synthesis within the highly customizable Minecraft environment for large-scale multi-task RL training, and we construct an efficient distributed RL framework to support this. Experimental results show RL significantly boosts interaction success rates by $4\times$ and enables zero-shot generalization of spatial reasoning across diverse environments, including real-world settings. Our findings underscore the immense potential of RL training in 3D simulated environments, especially those amenable to large-scale task generation, for significantly advancing visuomotor agents' spatial reasoning. 
\end{abstract}

\begin{links}
    \link{Code}{https://github.com/CraftJarvis/ROCKET-3}
\end{links}

\section{Introduction}

Reinforcement Learning (RL) has shown immense potential in solving complex tasks, particularly in sequential decision-making~\citep{dqn,alphago}. Generally, applying RL to train multi-task policies typically relies on meticulously designing reward functions from scratch to guide agents in learning specific task knowledge. However, this approach has been widely noted for problems like catastrophic forgetting~\citep{vithayathil2020survey} and multi-task interference~\citep{taylor2011introduction}, which severely hinder RL's generalization capabilities in complex multi-task environments. In recent years, the rapid advancement of Large Language Models (LLMs)~\citep{gpt4,DeepSeekAI2025DeepSeekR1IR} has introduced a fundamentally new paradigm for RL's application. It demonstrates that RL is no longer merely a tool for learning a specific task; instead, it can serve as a crucial technique during the post-training phase to enhance core LLM capabilities such as logical reasoning and instruction following. This paradigm shift in RL is largely attributable to two key factors: first, the general knowledge acquired during large-scale pre-training, and second, how ``next-token prediction'' unifies the LLM's task representation space, enabling the model to process diverse language tasks coherently. 

While RL has achieved remarkable success in language modeling, its triumph hasn't yet fully translated to visuomotor agents. A primary challenge lies in RL models' tendency to overfit specific tasks or environments, hindering the acquisition of generalizable behaviors and cross-environment generalization. This paper provides a preliminary answer to this challenge by demonstrating that RL-finetuned visuomotor agents can achieve zero-shot generalization of their enhanced spatial reasoning capabilities to unseen environments (including other 3D environments and the real world). 

To achieve this, we need to construct a unified and efficient multi-task goal space. We believe an ideal visuomotor agent's goal space should possess the following key properties: \textit{openness} to accommodate an infinite variety of tasks; \textit{unambiguity} to ensure the agent's precise understanding of task intent; \textit{scalability} to support large-scale task generation; and \textit{curriculum} property to enable the agent to progressively learn complex skills. After a thorough analysis of current mainstream task representation methods, we finally select \textit{cross-view goal specification}~\citep{cai2025rocket} as our unified task space. This means that any task involving interaction with a specific object in an open world can be uniformly represented by: selecting a novel camera view from which the target object is observable, and generating a precise segmentation mask of that target object. This representation inherently fuses visual information with task objectives, laying a solid foundation for subsequent RL training. 

To support large-scale RL post-training, we face the challenge of synthesizing training tasks at scale. We choose the highly customizable open-world environment Minecraft as the RL training platform for our policies. Minecraft's flexibility allows us to synthesize a vast number of task instances, spanning various visual perspectives and exhibiting smooth transitions in difficulty, by randomly sampling factors such as world seed, terrain, camera view, and target object. This automated task generation mechanism resolves the bottleneck of manual task design, enabling us to conduct large-scale multi-task training unprecedented in scope. 
To address the engineering challenges posed by large-scale RL training, we further implement an efficient distributed RL framework. This framework effectively overcomes the bottlenecks of trajectory collection and data transmission prevalent in existing RL frameworks~\citep{ray} within complex environments (like Minecraft), while also supporting stable training of long-sequence Transformer-based policies, ensuring we can leverage the synthesized large-scale tasks. 

Extensive RL post-training within the complex Minecraft on $\mathbf{100,000}$ tasks reveals a remarkable $4\times$ increase in the agent's success rate in performing interactions under significant variations in cross views. Notably, we further demonstrate the efficacy of this RL-enhanced agent by deploying it zero-shot to DMLab~\citep{dmlab}, Unreal Engine~\cite{unrealzoo}, and real-world settings, where we observe compelling evidence of its generalized cross-view spatial reasoning capabilities. These findings strongly validate that RL can serve as a potent post-training mechanism for substantially augmenting the core competencies of visuomotor policies, endowing them with exceptional domain generalization. 
\textbf{Our contributions are as three-fold:}
\begin{enumerate}
    \item We propose an innovative method for large-scale, automated synthesis, generating over $\mathbf{100,000}$ Minecraft training tasks to overcome the bottleneck of manual design. This enables us to perform the first multi-task reinforcement learning in the challenging Minecraft. 
    \item We develop an efficient distributed RL framework to address engineering challenges in complex environments, ensuring stable training of long-sequence policies. 
    \item We empirically demonstrate that RL can serve as a powerful post-training mechanism for visuomotor policies, showing a remarkable $4\times$ increase in interaction success rates and compelling zero-shot generalization of cross-view spatial reasoning in diverse, unseen environments. 
\end{enumerate}

\begin{table*}[htbp]
\footnotesize 
\centering
\caption{\textbf{Key Properties of Effective Task Spaces for Embodied Agents.}}
\label{tab:task_properties}
\begin{tabular}{p{0.10\textwidth}p{0.85\textwidth}}
\toprule
\textbf{Openness} & Refers to the diversity and infinitude of the task space. It enables agents to continuously encounter novel visual configurations, object arrangements, or interaction scenarios, preventing rote memorization. This ensures agents develop robust and generalizable visuomotor policies capable of handling unseen real-world complexities. \\
\midrule
\textbf{Unambiguity} & Ensures that each task instance has clear, well-defined metrics and verifiable success criteria. For visuomotor agents, this means the goal state or action execution must be precisely measurable. Such clarity is vital for expert demonstrations in imitation learning (IL) and for designing effective reward signals during reinforcement learning (RL) fine-tuning. \\
\midrule
\textbf{Scalability} & Emphasizes that the task space must facilitate the automated and large-scale generation of both demonstration data for IL pre-training and expanded task sets for RL fine-tuning. Crucially, reward functions for these tasks must be easily and efficiently designable, or verifiable without extensive human intervention. \\
\midrule
\textbf{Curriculum} & A task space with curriculum properties provides a smooth transition in difficulty, offering a progressive learning path from simple to complex. It contains a spectrum where agents gradually master basic skills, with simpler tasks serving as necessary building blocks for more intricate ones, thus facilitating knowledge transfer. \\
\bottomrule
\end{tabular}
\vspace{-1em}
\end{table*}

\section{Related Works and Preliminaries}

\paragraph{Imitation Learning} IL centers on enabling an agent to learn behavior policies by observing expert demonstrations. It transforms complex decision-making into a supervised learning task: given a state $S_t$, predict the action $A_t$ an expert would take. This is typically achieved by minimizing the behavioral discrepancy between the policy $\pi_\theta$ and the expert policy $\pi_E$, often using maximum likelihood estimation for discrete actions in behavior cloning~\citep{bc}:
\begin{equation}
\max_{\theta} \mathbb{E}_{(S_t, A_t) \sim D_E} \left[\log \pi_\theta(A_t|S_t)\right].
\end{equation}
Through large-scale expert data, IL empowers agents to acquire rich world knowledge, generalized patterns, and an implicit understanding of task intentions, rapidly building foundational behavioral capabilities. 
For instance, large language models (LLMs) are fundamentally driven by large-scale imitation learning via next token prediction, internalizing language structures and world knowledge from vast text corpora \citep{gpt2, gpt3}. Similarly, in visuomotor control, many leading vision-language-action models (VLAs), like DeepMind's RT-X series~\citep{rt-1, rt-2}, are pre-trained on massive robot demonstration datasets~\cite{rt-x} using IL, gaining an initial grasp of object physics, operational causality, and task instructions. 
However, IL's effectiveness is constrained by expert data quality, preventing it from surpassing expert performance or enabling autonomous exploration and error correction. Crucially, it's prone to the \textbf{covariate shift} problem~\citep{ross2011reduction}---where the agent's actions lead to states $S'$ unseen in expert data, causing performance to degrade sharply. 

\paragraph{Reinforcement Learning} With its capacity for exploration and learning from rewards, RL has achieved remarkable success in single-task, clearly defined domains, such as AlphaGo~\citep{alphago} for Go or MOBA games like Dota 2~\citep{ye2020mastering}. Unlike IL, RL inherently allows agents to explore beyond expert data, discover novel strategies, and self-correct through environmental feedback, thus overcoming the covariate shift problem and even surpassing expert performance. The core optimization objective in RL is to maximize the agent's expected cumulative reward:
\begin{equation}
 \max_{\theta} \mathbb{E}_{\tau \sim \pi_\theta} \left[ \sum\nolimits_{t=0}^T \gamma^t R_t \right].
\end{equation}
However, attempts to apply this RL paradigm for training general-purpose agents in multitask, open-world, or high-dimensional observation spaces have frequently failed~\citep{electronics9091363}. This is mainly because, in complex open-world scenarios, RL faces significant challenges, notably sample inefficiency and sparse reward signal~\citep{minedojo, vpt, worldmc}. It is incredibly difficult to construct a dense reward signal that universally incentivizes behavior across multiple tasks. This often leads to agents struggling to receive effective feedback during exploration, and they can easily fall into the traps of catastrophic forgetting and negative transfer, causing them to unlearn previously acquired skills or for different tasks to conflict. A deeper underlying reason is that \textit{pure RL lacks prior general world knowledge and common sense}, forcing the agent to learn everything about the environment and tasks from scratch, which is highly inefficient in complex, open-ended settings. 

\paragraph{Foundation-to-Finesse Learning}
Given the complementary strengths of IL (efficient knowledge acquisition) and RL (exploration and refinement), and acknowledging the limitations of pure IL in generalization and the sample inefficiency of training RL from scratch in multi-task scenarios, the prevailing paradigm for LLM training has evolved into an effective combination of both~\citep{ouyang2022training, DeepSeekAI2025DeepSeekR1IR}. This approach features a clear, progressive training flow designed to build powerful agents~\citep{ze2023visual, ptgm}. 
First, IL serves as the \textit{builder of foundational knowledge and implicit reasoning capabilities}. By training on vast amounts of expert data, agents efficiently learn and internalize large-scale general world knowledge, common sense, behavioral patterns, and an implicit understanding of diverse tasks. This observation-acquired generalization lays the groundwork for subsequent causal and spatial reasoning, enabling agents to comprehend various instructions and contexts and produce initial, expected responses. 
Subsequently, RL takes on the crucial role of \textit{refining and applying explicit reasoning capabilities}. Building upon the solid foundation laid by IL, agents enter real or simulated environments to further optimize their policy through active trial-and-error and reward feedback. At this stage, RL is no longer blind exploration from scratch but rather fine-tuning based on a well-initialized policy. 
This progressive relationship allows agents to efficiently learn ``how to do'' from imitation, and then ``how to do better'' through RL, ultimately translating implicit knowledge into actionable, verifiable reasoning capabilities. 

\section{Task Space for Generalizable RL}
In traditional multi-task RL, a visuomotor agent learns to master a small set of k predefined tasks. The task representation in this paradigm is often a simple identifier (e.g., a one-hot vector), which lacks the semantic structure required for meaningful knowledge transfer, thus hindering generalization. 
Our objective is more ambitious: to enable a policy to generalize from $k$ training tasks to $n \gg k$ novel tasks, or even to entirely new 3D environments. Achieving this leap requires a unified task space that can seamlessly bridge training and generalization. We argue that an ideal task space must inherently satisfy four properties, shown in Table~\ref{tab:task_properties}. Next, we analyze the following common task spaces. 

\begin{figure*}[htbp]
\begin{center}
\includegraphics[width=0.99\linewidth]{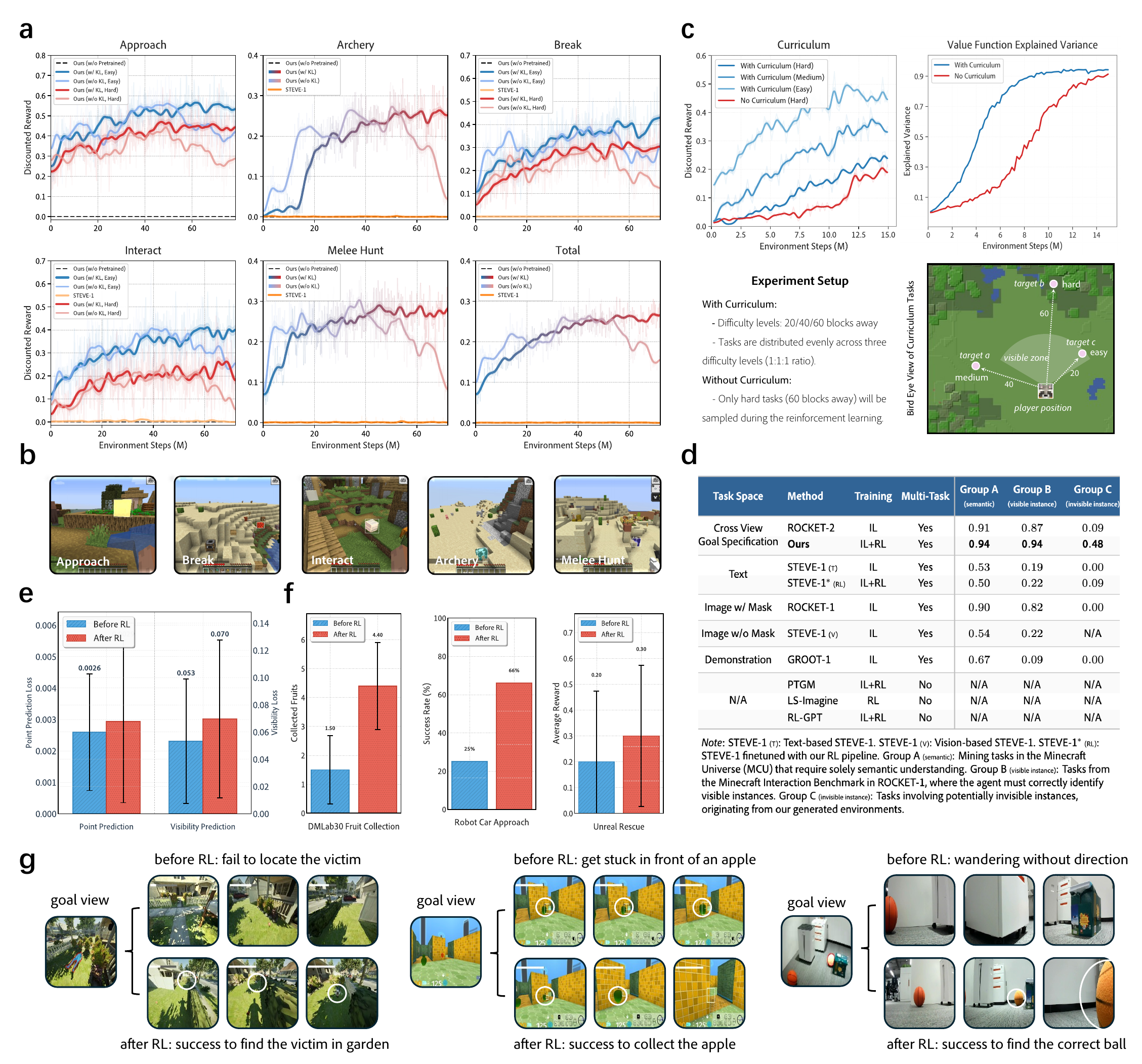}
\end{center}
\caption{
\textbf{RL Post-Training Boosts Generalizable Spatial Reasoning and Open-World Interaction Capabilities.} 
\textbf{(a)} RL training curves for five skills in the Minecraft environment. This panel shows simultaneous performance gains across all skills. It also highlights the policy's performance collapse in later training stages without a KL divergence constraint.
\textbf{(b)} Sample target viewpoints for each skill during training, encompassing various camera view ranges (e.g., eye-level and top-down). ``Archery'' involves long-range interaction with mobs, while ``Melee Hunt'' requires close-quarters combat. 
\textbf{(c)} Comparison of curriculum-based training (mixed difficulties) with non-curriculum training (hard tasks only). The ``Discounted Reward'' plot on the left shows curriculum learning leads to higher training efficiency and faster reward accumulation, while the ``Value Function Explained Variance'' plot on the right demonstrates it also accelerates value function learning. 
\textbf{(d)} Results table for current SOTA goal-conditioned agents in Minecraft. Success rate is reported. Our agent is the first to achieve successful multi-task RL in challenging Minecraft environment. Several representative single-task RL agents are also listed for reference.
\textbf{(e)} Point Prediction and Visibility Prediction loss comparison before and after RL training. Losses for these heads on the pre-training dataset remain largely unchanged despite not being optimized during RL, indicating that RL preserved the policy's original representations. 
\textbf{(f)} This panel shows significant improvements in DMLab30 fruit collection, robot car approach, and Unreal rescue reward after RL training, demonstrating the model's effective generalization to unseen 3D worlds. 
\textbf{(g)} Case studies of domain transfer. We analyze some successful and failure cases here. More details can be found in supplementary details. We performed 32 runs for each experiment. 
}
\label{fig:exp}
\end{figure*}

\noindent \textbf{Natural Language} as a task space offers high \textit{openness} due to its inherent expressiveness and compositionality, easily facilitating diverse task sets with varying \textit{curricula} difficulties. However, it exhibits high \textit{ambiguity} for fine-grained spatial relationships, complicating large-scale reward design and verification, thus limiting its \textit{scalability} for precise localization and manipulation tasks. When the target object is invisible, language no longer provides meaningful guidance for exploration. Figure~\ref{fig:exp}\textbf{a} illustrates the failure of the language space in multi-task RL within complex Minecraft. 

\noindent \textbf{Instance Image} defines tasks by providing close-up photos of a target object, often requiring the object to dominate the frame (e.g. $70\%$ coverage)~\citep{krantz2023navigating}. Although semantically rich, this representation inherently deemphasizes spatial context, limiting its utility for complex spatial reasoning tasks. Lacking an explicit instance cue, this method suffers from target ambiguity, especially in the presence of other small objects in the background. 
And, it struggles with \textit{openness} and \textit{curriculum} due to a narrow range of possible visual contexts, and its focus on appearance matching rather than understanding spatial relationships. 

\noindent \textbf{Cross-View Goal Specification (CVGS)} offers a method to specify any goal object using a segmentation mask from a third-person view. This approach inherently overcomes the rigid ``qualification'' constraints of \textit{Instance Image} and, more importantly, demands the agent to reason about spatial relationships between its current view and the third-person goal view. Its flexibility allows precise control over task difficulty by adjusting view distance and overlap, making it strong in \textit{openness} and \textit{curriculum}. Its clear definition of goals also ensures high \textit{unambiguity} and efficient \textit{scalability} for large-scale task generation and reward verification. An notable advantage is: \textit{even if the agent can't directly see the target object, the landmark shared across the views can still offer crucial guiding information.} 
\textbf{We adopt CVGS as our goal space because it naturally facilitates cross-domain generalization. The core capabilities it requires, reasoning about visual views and spatial information within the same domain, are inherently suited for this.}

\begin{figure*}[htbp]
\begin{center}
\includegraphics[width=0.99\linewidth]{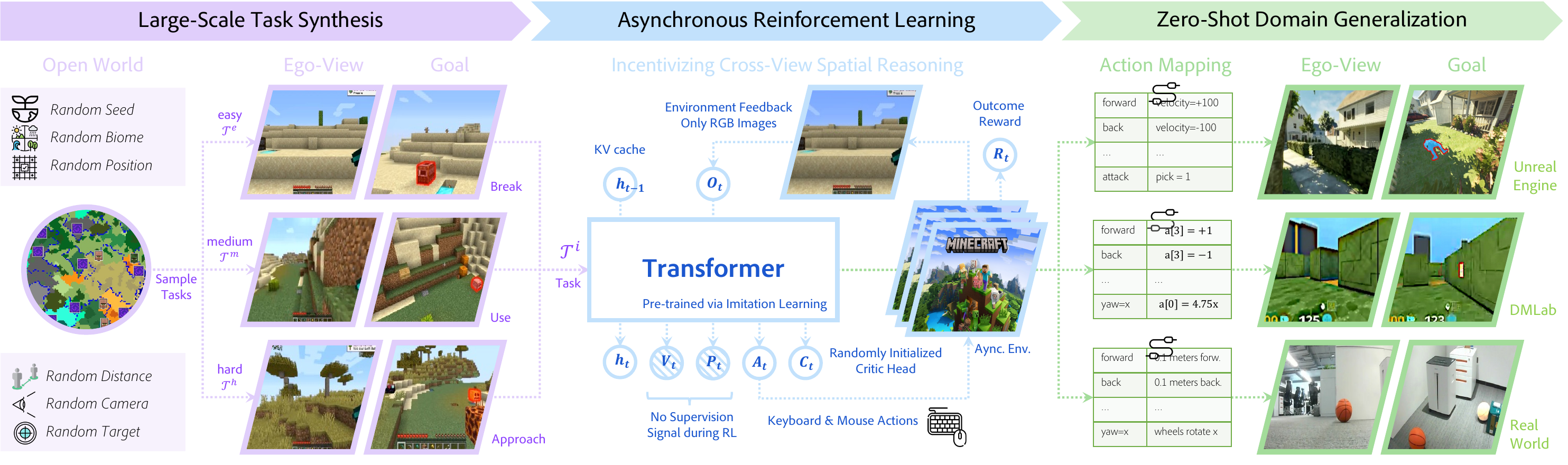}
\end{center}
\caption{
\textbf{The Post-Training Pipeline.} We synthesize large-scale, mixed-difficulty cross-view interaction tasks in an open-world environment by randomly sampling terrain, distances, target objects, and camera views. The foundational policy is fine-tuned using our distributed RL framework and then deployed in unseen 3D worlds via a simple action space mapping. 
}
\label{fig:pipeline}
\end{figure*}

\section{Pipeline Design} 
\paragraph{Task Formulation} We define a task instance $\mathcal{T}$ using a combination of pixel images and instance mask: $\mathcal{T} = \left<O_1, O_g, M_g, E \right>$, where $O_1$ is the initial agent view obtained by resetting the environment, $O_g$ is the goal observation, provided from a distinct, often human-centric or third-person viewpoint. Crucially, $O_g$ includes a precise segmentation mask $M_g$ that explicitly highlights the target object. $E$ denotes the interaction event, e.g. \textit{break item}, \textit{use item}, \textit{pick up} and \textit{place}. 
The agent's policy, denoted as $\pi_\theta(A_t | O_{1:t}, \mathcal{T})$, is a network that maps the agent's observations $O_{1:t}$ and the task instance $\mathcal{T}$ to a distribution over actions $A_t$. The core challenge is to learn a cross-view alignment; that is, to understand the spatial relationship between its own $O_{1:t}$ and the goal specified by $M_g$ in $O_g$. 

\paragraph{Pre-Training via Imitation Learning} 
Our IL stage follows~\cite{cai2025rocket}, which pre-trains policies on large-scale trajectories collected via backward trajectory relabeling. We formulate the dataset with $N$ trajectories as: 
\begin{equation}
    \mathcal{D} = \{(O^{(i)}_{1:T}, A^{(i)}_{1:T}, M^{(i)}_{1:T}, V^{(i)}_{1:T}, P^{(i)}_{1:T}, E^{(i)}) \}_{i=1}^{N}, 
\end{equation}
where $V^{(i)}_{t}$, $P^{(i)}_{t}$, and $M^{(i)}_{t}$ respectively denote the target object's visibility, geometric centroid, instance mask in frame $O^{(i)}_{t}$. As the target object remains the same with each trajectory $\tau_i$, we can sample any frame index $g \in [1,T]$ to build a task instance $\mathcal{T}^{(i)} = \left<O^{(i)}_1, O^{(i)}_g, M^{(i)}_g, E^{(i)} \right>$. 
To enhance the model's sensitivity towards target perception, we maximize the log-likelihood of a joint distribution as objective:
\begin{equation}
\max_\theta \frac{1}{NT} \sum_{i=1}^{N} \sum_{t=1}^{T} \log \pi_\theta(A^{(i)}_t, V^{(i)}_t, P^{(i)}_t | O^{(i)}_{1:t}, \mathcal{T}^{(i)}). 
\label{eq:il_objective}
\end{equation}

\paragraph{Large-Scale Cross-View Task Synthesis} 
Given any task $\mathcal{T} = \left<O_1, O_g, M_g, E \right>$, cross-view spatial reasoning involves analyzing the relationship between history views $O_{1:t}$ and goal view $O_g$ to implicitly plan an executive path. Therefore, the discrepancy between $O_1$ and $O_g$ naturally characterizes the task's difficulty, with difficulty changes exhibiting a smooth, continuous relationship. We observe that pre-trained agents show \textbf{weak} foundational spatial reasoning, succeeding only when $O_1$ and $O_g$ are minimally different. 
\textit{We aim to explore if RL can enhance this spatial reasoning ability and enable transfer to other 3D environments.} 

To this end, we designed an automated task synthesis method based on the Minecraft environment. Specifically, we first randomly sample a spawn location $p_0$ in the world and generate interactive objects (e.g., blocks, mobs) in its vicinity. Subsequently, we sample a distance $d$ (which directly influences task difficulty), teleport the player to a location at that distance, and adjust the camera view to encompass at least one object, thus obtaining a novel goal view $O_g$. 
We access the voxel information around the player in the Minecraft simulator, then select one of these objects as the interaction target. The \textit{bottom-center coordinate} of this object is $G=(G_x, G_y, G_z)$. By combining this with the player's \textit{eye center coordinate} $U=(U_x, U_y, U_z)$, and the player's \textit{yaw angle} $\theta_y$ and \textit{pitch angle} $\theta_p$, we can construct the corresponding rotation matrix
\begin{align}
    \begin{adjustbox}{width=0.9\linewidth}
        $R_M =
        \begin{bmatrix}
        \cos(\theta_{y}) & 0 & \sin(\theta_{y}) \\
        -\sin(\theta_{p})\sin(\theta_{y}) & \cos(\theta_{p}) & \sin(\theta_{p})\cos(\theta_{y}) \\
        -\cos(\theta_{p})\sin(\theta_{y}) & -\sin(\theta_{p}) & \cos(\theta_{p})\cos(\theta_{y})
        \end{bmatrix}$, 
    \end{adjustbox}
\end{align}
Therefore, the object in the camera coordinate system can be expressed as $C = R_M(G - U)$. 
Subsequently, based on the dimensions of the $O_g$ screen $H \cdot W$, the vertical field of view angle $f_y$, and the principles of perspective projection, we can calculate its values in \textit{ normalized device coordinates (NDC)} $(n_x, n_y)$, which are then finally converted into the screen's \textit{pixel coordinates} $(u, v)$
\begin{gather}
    f_x = 2 \cdot \arctan\left(\tan\left(f_y/2\right) \cdot W/H\right), \\
    n_x = \frac{C_x}{C_z \cdot\tan(f_x/2)}, n_y = \frac{C_y}{C_z \cdot \tan(f_y/2)}, \\ 
    u = (n_x + 1)/2\cdot W, v = (1-n_y)/2 \cdot H.
\end{gather} 
As individual voxels cannot precisely represent an object's complete shape, we incorporate a \textit{Segment Anything Model (SAM)}~\citep{sam2}. This model utilizes a series of sampled points from the voxel's cube as prompts to extract the target object's full mask $M_g$ in pixel space. 
After generating the cross view, we use the ``\texttt{spreadplayers} $p_0$ \texttt{distance}'' command to generate starting position and $O_1$. 
The \texttt{distance} parameter directly influences task difficulty and curriculum design.
Rewards are then automatically generated by detecting changes in the object's voxels within the simulator, leading to an outcome reward. 

\begin{table*}[htbp]
\footnotesize 
\centering
\caption{\textbf{Overview of Training and Testing Environments.}}
\label{tab:environments}
\begin{tabular}{p{0.14\textwidth}p{0.81\textwidth}} 
\toprule
\textbf{Minecraft} \newline \citep{minerl} & \textbf{Version:} 1.16.5. \textbf{Observations:} $640 \times 360$ pixels, 70-degree FoV. \textbf{Actions:} Mouse and keyboard operations. \textbf{Purpose:} Primary training and testing platform; rich dataset for pre-training, high freedom for large-scale task synthesis, crucial for studying cross-view spatial reasoning and open-world interaction. \\
\midrule
\textbf{Unreal} \newline \citep{unrealzoo} & \textbf{Observations:} $640 \times 480$ pixels, highly realistic textures, visually complex. \textbf{Actions:} Movement, view adjustment, jumping, interaction. \textbf{Purpose:} Dedicated testing platform for personnel search and rescue missions, assessing agent's ability to locate and transport casualties using cross-view clues in a high-fidelity environment. \\
\midrule
\textbf{DMLab} \newline \citep{dmlab} & \textbf{Observations:} $320 \times 240$ visual images. \textbf{Actions:} Comparable to Minecraft (Movement, view adjust, shoot, ...). \textbf{Purpose:} Game-based assessment of embodied agents' navigation and interaction skills within partially observable settings (e.g., fruit collection). Utilized for validating generalization capabilities. \\
\midrule
\textbf{Real World} \newline \citep{mecanum} & \textbf{Physical Embodiment:} Robot car with Mecanum wheels. \textbf{Observations:} $640 \times 360$ pixels from a 110-degree camera. \textbf{Purpose:} To ascertain whether learned cross-view spatial reasoning capabilities generalize to real-world. \\
\bottomrule
\end{tabular}
\end{table*}

\paragraph{Post-Training via Reinforcement Learning} 
We optimize the policy using a combination of the Proximal Policy Optimization (PPO)~\citep{ppo} and a KL constraint $\mathcal{L} = \mathcal{L}^\text{PPO} + \beta \cdot \mathcal{L}^\text{KL}$. Minimizing the KL divergence could enhance PPO's training stability by preserving knowledge from a reference policy $\pi_\text{ref}$, where $\pi_\text{ref}$ is the initial pre-trained policy: 
\begin{equation}
    \mathcal{L}_\text{KL} = D_\text{KL}\left( \pi_\theta(\cdot|O_{1:t}, \mathcal{T}) \parallel \pi_{\text{ref}}(\cdot|O_{1:t}, \mathcal{T})\right). 
\end{equation}
The policy loss of standard PPO is formulated as follows:
\begin{align}
    \begin{adjustbox}{width=0.88\linewidth}
        $\displaystyle \mathcal{L}^\text{PPO} = -\mathbb{E}_t \left[ \min\left( r_t(\theta) \hat{A}_t, \text{clip}(r_t(\theta), 1-\epsilon, 1+\epsilon) \hat{A}_t \right) \right]$, 
    \end{adjustbox}
\end{align}
where $\hat{A}_t$ is the generalized advantage function \citep{gae}, $r_t(\theta) = \pi_\theta(\cdot|O_{1:t}, \mathcal{T})/\pi_{\theta_{\text{old}}}(\cdot|O_{1:t}, \mathcal{T})$ is the importance sampling ratio. During the RL process, we only optimized the action head, while omitting supervision for the auxiliary heads that predict object visibility $V_t$ and the centroid point $P_t$. 
Interestingly, our experiments show that even without explicit supervision, the policy retains the functionality of these two heads after RL post-training, suggesting that the spatial reasoning learned for action control implicitly benefits these perceptual tasks.

\begin{figure}[t]
\begin{center}
\includegraphics[width=0.99\linewidth]{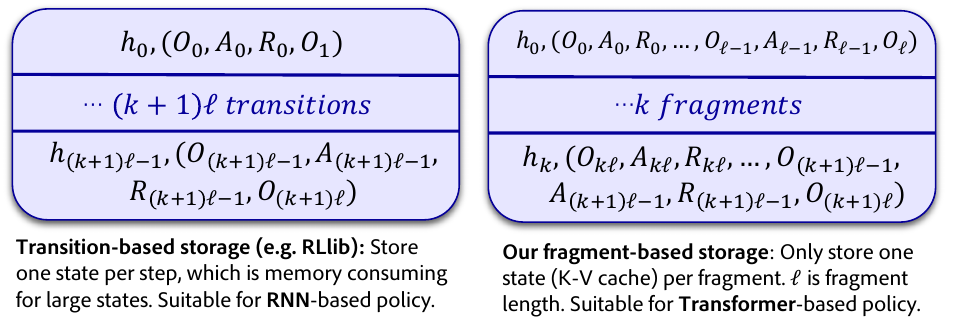}
\end{center}
\caption{
\textbf{Trajectory Storage Comparison}. 
}
\label{fig:rl_storage}
\end{figure}

\paragraph{Distributed RL Framework Design}
No off-the-shelf RL framework currently meets our specific needs, primarily due to the following considerations: \textit{high communication costs}, \textit{simulator instability}, and \textit{long-term dependency handling}. 
To tackle these issues, our framework assumes a cluster composed of a shared Network Attached Storage (NAS) and multiple compute nodes, incorporating the following core mechanisms: 
\textbf{Asynchronous Data Collection}: Rollout workers can be deployed on any compute node. Each worker comprises an inference model and N independent Minecraft instances. These instances asynchronously send requests to a queue, and the model performs batch inference when the queue reaches its specified batch size. \textbf{Optimized Data Transfer}: We use Ray~\citep{ray} to organize different compute nodes into a cluster. However, the trajectories collected by rollout workers are not sent directly to the trainer. Instead, they are stored directly in a database on the shared NAS, with the trainer receiving only data indices. This strategy significantly alleviates the consumption of network bandwidth during training, addressing the shortcomings observed in modern frameworks like RLlib~\citep{rllib}. 
\textbf{Support for Long Sequence Training}: 
To facilitate the training of our Transformer-based policy on long sequences, we introduce a memory-efficient, fragment-based storage method. Unlike traditional transition-based storage, our approach stores the K-V cache state (about $10$ MB per step) only once per fragment (as shown in Figure~\ref{fig:rl_storage}), drastically reducing memory overhead. This, coupled with truncated Backpropagation Through Time (tBPTT), allows the policy to leverage K-V cache from thousands of prior frames ($O_{1:t-1}$), which is vital for capturing long-term dependencies in hard tasks. 
Our framework allows us to simultaneously launch 72 Minecraft instances in 3 compute nodes, achieving a collection speed of about $1000$ FPS. We will open-source our RL training framework to foster further RL research in complex environments. Details about the RL framework can be found in supplementary materials. 

\section{Experiments}

\subsection{Environments}

As shown in Table \ref{tab:environments}, we focus on environments where the observation space is purely pixel-based and the action space is abstractly defined to facilitate generalization. This action space encompasses omnidirectional movement control (akin to WASD key combinations), continuous camera view adjustments (similar to mouse-controlled pitch and yaw), and a set of functional object interaction actions (e.g., picking up or dropping items). This unified design aims to enable the effective transfer of learned capabilities across diverse, complex environments. \textit{For all the 3D worlds, the visual inputs are resized to $224 \times 224$ before feeding into the agent. }

\subsection{RL Post-Training Discoveries in Minecraft}
We conduct post-training on about $\mathbf{100,000}$ sampled tasks within the Minecraft environment. These tasks encompass various interaction types, including \textit{Approach}, \textit{Break}, and \textit{Interact}, as well as \textit{Hunt} (subdivided into \textit{Melee Hunt} and \textit{Archery}). Examples are shown in Figure \ref{fig:exp}\textbf{b}. 
To facilitate curriculum learning for RL training, we implement difficulty levels for \textit{Approach}, \textit{Break}, and \textit{Interact} tasks. In \textit{easy} difficulty tasks, the Manhattan distance between the agent's starting position and the target location was approximately $20$ blocks. Conversely, \textit{hard} difficulty tasks extended this distance to roughly $60$ blocks. By analyzing the training curves in Figure~\ref{fig:exp}, we observe the followings. 

\paragraph{$4\times$ Performance Leap Under Complex Views}
Task performance across all interaction types significantly improved, with the average success rate increasing from $7\%$ to $28\%$. Notably, for \textit{Archery}, the success rate surged from less than $1\%$ after pre-training to $28\%$ following RL post-training, indicating that RL can unleash rare capabilities from pre-training. The improved success rates on \textit{hard} tasks further demonstrate the model is acquiring exploration abilities. 

\paragraph{Ensuring Stable RL Post-Training with KL}
Figure \ref{fig:exp}\textbf{a} reveals KL divergence is key to RL post-training stability. Specifically, this KL divergence is computed with respect to the initial imitation learning pretrained policy. Models with KL divergence (w/ KL) show more stable learning and consistently higher performance, avoiding the fluctuations and collapse seen in models without it (w/o KL). We also find that policies without pre-training failed in multi-task RL, highlighting Minecraft's complexity and tasks' difficulty. 

\paragraph{Language-Based RL: STEVE-1's Adaptation Bottleneck}
Figure \ref{fig:exp}\textbf{a} shows language-based STEVE-1~\citep{steve1}, which is pre-trained on the Minecraft contractor data via imitation learning and post-trained with our RL pipeline, consistently achieves \textbf{near-zero performance} during RL stage. This highlights a critical limitation: \textit{natural language inherently struggles to support effective spatial context reasoning for distant or occluded target objects}. Conversely, in situations where objects are not visible, our method can leverage background and landmark objects from a third-view perspective to aid in spatial reasoning, thereby providing effective exploration guidance for RL. 

\paragraph{Mixed-Difficulty Curriculum for Accelerated Learning}
Unlike traditional easy-to-hard progressions, our curriculum adopts a mixed-difficulty training strategy. As Figure \ref{fig:exp}\textbf{c} shows, using the \textit{Break} interaction, we define three distinct difficulty levels: \textit{Easy}, \textit{Medium}, and \textit{Hard}, which are characterized by Manhattan distances of $20$, $40$, and $60$ blocks, respectively. In our curriculum setup, the model is trained simultaneously and uniformly across all three difficulty levels. In contrast, the non-curriculum baseline is trained exclusively on the \textit{hard} task. 
Notably, even though \textit{hard} tasks constitute only one-third of the sampling frequency in our curriculum setting, we observe a higher performance improvement. The \textit{explained variance} curve further illustrates this: the curriculum-trained model (blue) converges faster and reaches higher explained variance than the non-curriculum baseline (red). This strongly demonstrates that a mixed-difficulty curriculum can substantially accelerate the learning of complex skills in RL environments. 

\paragraph{Robustness of Intrinsic Spatial Reasoning} 
Figure \ref{fig:exp}\textbf{e} reveals a key discovery: auxiliary prediction heads (centroid and visibility, Equation \ref{eq:il_objective}) maintain strong performance post-RL, degrading only slightly despite no explicit training during this stage. This sustained performance demonstrates the robustness of the agent's intrinsic spatial reasoning. It indicates that the fundamental spatial understanding, fostered by the cross-view goal alignment task space, persists largely unchanged, preventing overfitting to downstream objectives.

\subsection{Baselines Comparison in Minecraft}

To benchmark our model in complex Minecraft interactions, we compare it against mainstream end-to-end baselines: STEVE-1, ROCKET-1~\citep{cai2024rocket}, ROCKET-2~\cite{cai2025rocket}, GROOT~\cite{groot-1}, PTGM~\citep{ptgm}, RL-GPT~\citep{liu2024rlgpt} and LS-Imagine~\citep{ls-imagine}. Given the significant variations among these baselines in terms of single/multi-task focus, task space, and training methods, we construct three progressively challenging task groups: \textit{semantic understanding}, \textit{visible instance interaction}, and \textit{invisible instance interaction}. The \textit{semantic} group includes tasks like ``chop tree'' and ``hunt sheep with arrow'', completed upon semantic match. The \textit{visible instance} group requires interaction with a specific object visible to the agent. The \textit{invisible instance} group utilizes a third-view to specify the target, as it's otherwise not visible from the agent's current perspective. All three task groups necessitate multi-task capabilities, rendering many existing RL-based baselines (e.g. PTGM, RL-GPT) unsuitable due to their single-task nature. Figure \ref{fig:exp}\textbf{d} illustrates that most baselines achieve success rates only in the first two task groups, whereas our proposed method uniquely attains a $\mathbf{48}\%$ success rate in the third, most challenging group. This clearly demonstrates our approach's significant superiority over existing baselines in handling complex, target-invisible Minecraft interaction tasks.

\subsection{Generalizing RL Results Beyond Minecraft}

To validate our method's generality, we investigate RL-enhanced capabilities transferring to unseen 3D worlds. We experiment in DMLab, Unreal virtual environments, and with a real-world Mecanum-wheeled robot. These share a pixel observation space and an action space abstractable to omnidirectional movement, camera adjustment, and functional presses, enabling efficient policy adaptation via simple mapping. We present detailed adaptation, tasks, and view selection in the supplementary materials. 

Figures \ref{fig:exp}\textbf{f} and \textbf{g} present quantitative results and case studies. We observe the pre-trained policy shows weak generalization: success rates are low even with minor $O_1$ -- $O_g$ differences (e.g., $O_g$ eye-level, target visible in both). This baseline generalization stems from the DINO pre-trained ViT backbone seeing diverse 3D textures. However, the RL-enhanced policy significantly improves generalization: it succeeds even when $O_g$ presents a bird's-eye view and the target is invisible in $O_1$. Notably, in the real-world ball-finding task, RL boosts success by up to $\mathbf{41}\%$, highlighting its substantial practical potential. Nevertheless, we observe failures, such as the robot car sometimes hitting obstacles and frequently failing on medium-to-long-range approach tasks in the real world. This indicates the policy's performance is still impacted by the visual texture gap, underscoring the need for scaling up training worlds. 

\section{Conclusion}
This work validates that reinforcement learning significantly boosts visuomotor agents' cross-view reasoning and interaction skills. We show these enhanced abilities generalize across diverse 3D environments, including the real world. We've also gained valuable insights from the RL post-training process. Future work will explore unified RL training for 3D worlds with varied action spaces.

\bibliography{refs}

\clearpage

\appendix

\twocolumn[{
\begin{center}
    \LARGE \textbf{Supplementary Materials} \\[0.5ex]
    \large Implementation Details and Extended Results
\end{center}
\vspace{1.5em}
}]

\begin{table*}[t]
\centering
\renewcommand{\arraystretch}{0.95}
\setlength{\tabcolsep}{4pt}
\caption{Key hyperparameters for PPO training.}
\label{tab:hyperparameters_compact}
\begin{tabular}{@{} l r @{\hspace{8mm}} l r @{}}
\toprule
\textbf{Hyperparameter} & \textbf{Value} & \textbf{Hyperparameter} & \textbf{Value} \\
\midrule
\multicolumn{4}{c}{\textit{PPO Algorithm}} \\
\cmidrule(r){1-4}
Learning Rate & $2\times 10^{-5}$ & Weight Decay & 0.04 \\
Discount Factor ($\gamma$)& 0.999 & Max Gradient Norm & 0.5 \\
GAE Lambda ($\lambda$) & 0.95 & Log Ratio Range & 1.03 \\
PPO Clip Ratio & 0.2 & KL Divergence Coeff. ($\rho$) & 0.2 \\
Value Function Coeff. & 0.5 & KL Coeff. Decay & 0.9995 \\
\midrule
\multicolumn{4}{c}{\textit{Training Configuration}} \\
\cmidrule(r){1-4}
Context Length & 128 & Training Iterations & 4000 \\
Effective Batch Size & 10 & Fragment Length & 256 \\
Epochs per Iteration & 1 & Automatic Mixed Precision & True \\
\midrule
\multicolumn{4}{c}{\textit{Replay Buffer}} \\
\cmidrule(r){1-4}
Max Chunks & 4800 & Fragments per Chunk & 1 \\
Max Reuse & 1 & Max Staleness & 1 \\
\bottomrule
\end{tabular}
\end{table*}

\section{Cross-View Task Synthesis Details}

To generate a task, defined as $\langle O_1, O_g, M_g, E \rangle$, we follow a structured procedure:

First, a world seed is sampled, and the player is randomly teleported to an available location ($p_0$) within a randomly selected biome. Subsequently, an interaction type ($E$) is chosen from a predefined set: \textit{Approach}, \textit{Break}, \textit{Interact}, and \textit{Hunt}. Corresponding entities (e.g., blocks and mobs) relevant to the chosen interaction type are then generated within a predefined entity range around $p_0$.

Next, a random view range is determined by $\Delta x,\Delta y,\Delta z$ coordinates, along with pitch and yaw angles. The player is then teleported to the resulting cross-view position ($p_g$) to obtain the cross-view observation ($O_g$). Leveraging voxel information from the Minecraft simulator, a target object is selected from the generated entities, provided it is visible from $p_g$. 
The centroid point and bounding box of this target object within $O_g$ are extracted. These serve as prompts for SAM2 (using its largest checkpoint for optimal results) to generate the target's mask ($M_g$).

The initial observation ($O_1$) is generated using the command \texttt{/spreadplayers} around $p_0$ within a selected distance. The \texttt{distance} is randomly selected from $\{20,40\}$, with each value representing a different level of task difficulty.

To enhance task diversity, an alternative entity generation method is occasionally employed. Instead of generating entities at $p_0$, entities are generated directly at $p_g$ by randomly sampling an unoccluded voxel. This approach is particularly beneficial for long-horizon tasks and certain edge cases within \textit{Interact} tasks.

The agent's inventory and armor are randomly generated, while ensuring that all pre-requirements for interacting with specific entities are met. For example, an \textit{Archery} task provides a bow and $64$ arrows, while a \textit{Melee Hunt} task equips the player with a random sword. Our reward design is intentionally simple, providing a binary reward for each task based on the return information supplied by the simulator.

\section{Reinforcement Learning Design}

\subsection{Training Details}

\paragraph{Model Choice}
For the Cross-View Goal Alignment task space, we utilize the $0.3$B pre-trained ROCKET-2 checkpoint. For the language task space, the $0.6$B STEVE-1 checkpoint is employed. During RL training, both the vision backbone of ROCKET-2 and the text encoder of STEVE-1 remain fixed. Prompts for STEVE-1 are selected from its established prompt lists. Notably, the \textit{Approach} task is not trained for STEVE-1, as it was not pre-trained for this specific objective.

\paragraph{Hyperparameter Settings}

We present the hyperparameter settings in Table \ref{tab:hyperparameters_compact}. For the original PPO without KL, $\rho$ is set to $0$, while the other parameters remain unchanged. To ensure training stability, we apply clipping to both the gradients and the log ratio.

\subsection{RL Framework Pipeline}

\begin{algorithm*}[h]
\caption{Core Logic of the Distributed RL Framework}
\label{alg:framework}
\begin{algorithmic}[1]
    \Procedure{RolloutWorker}{}
        \State \textbf{Initialize}: $N$ parallel environments, local model, buffer $B$
        \Loop
            \State Asynchronously collect observations $O = \{o_1, \dots, o_m\}$ from environments
            \If{inference queue is full}
                \State $A \gets \text{model.inference}(O_{\text{batch}})$ \Comment{Batched inference for GPU efficiency}
                \State Dispatch actions $A$ to corresponding environments
            \EndIf
            \State Store fragmemts $\{h_i, (s_{i+\ell}, a_{i+\ell}, r_{i+\ell}, s_{i+\ell+1})\}$ in local buffer $B$ \Comment{$h_i$ is the hidden state, $\ell$ is the fragment length.}
            \If{$B$ reaches threshold}
                \State Write fragment data from $B$ to NAS
                \State Append metadata of $B$ to index file on NAS
                \State Clear $B$
            \EndIf
        \EndLoop
    \EndProcedure
    \Statex
    \Procedure{Trainer}{}
        \State \textbf{Initialize}: Policy model $\pi_\theta$, optimizer
        \Loop
            \State Poll index file on NAS to find new trajectory indices
            \State Sample batch of long sequences $S$ from NAS using indices
            \State Initialize hidden state $h_0$
            \For{each truncated segment $S_k$ in $S$}
                \State $\mathcal{L}(\theta) \gets \text{calculate\_loss}(S_k, h_{k-1})$
                \State Calculate $\nabla_\theta \mathcal{L}(\theta)$ \Comment{Perform tBPTT}
                \State $h_k \gets h_{k-1}.\text{detach}()$ \Comment{Propagate hidden state for next segment}
            \EndFor
            \State Update model weights $\theta$ using aggregated gradients
            \State Periodically save model checkpoint $\theta$ to NAS
        \EndLoop
    \EndProcedure
\end{algorithmic}
\end{algorithm*}

\begin{figure*}
    \centering
    \includegraphics[width=0.99\linewidth]{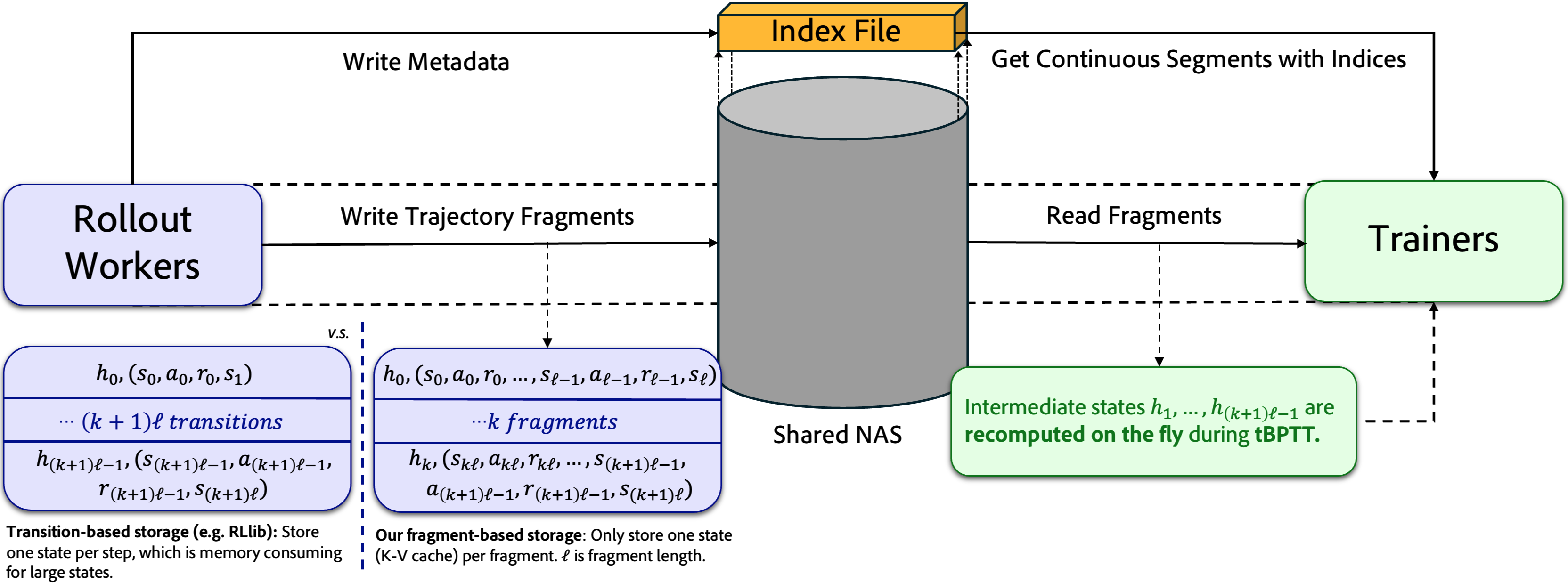}
    \caption{\textbf{Our fragment-based storage strategy.} Our rollout workers only save the initial latent states (K-V caches) at the beginning of each contiguous fragment. Latent states within the fragments are computed on the fly during tBPTT.}
    \label{fig:storage_pipeline}
\end{figure*}

Our distributed Reinforcement Learning framework is engineered to tackle large-scale, long-horizon training tasks within Minecraft. It operates on a compute cluster with a shared Network-Attached Storage (NAS) and leverages Ray for resource coordination and fault tolerance. The core design ensures scalability by decoupling data collection from training and optimizing inter-process communication.

The logic is split between two primary components: Rollout Workers for data collection and Trainers for model optimization. To minimize network overhead, workers write trajectory data (fragments) directly to the shared NAS. Synchronization is achieved using a lightweight index file containing only metadata, which trainers poll to discover new data. 

A key aspect of our design is a fragment-based storage strategy that optimizes for storage efficiency (Figure \ref{fig:storage_pipeline}). Unlike frameworks like RLlib~\citep{rllib} that store model-dependent latent states (K-V caches for Transformer-XL based models, which are disk space consuming) with each transition, our Rollout Workers only store the initial latent state \textbf{at the beginning of each contiguous fragment of experience}. This approach dramatically reduces the storage footprint, as a single latent state is stored for hundreds of transitions. The subsequent latent states within the fragment are then recomputed \textbf{on-the-fly} segment by segment during the Truncated Backpropagation Through Time (tBPTT) process, trading a small amount of computation for a massive reduction in disk space usage.

The trainer is specifically designed to support long-sequence policy training for stateful models. It samples long, overlapping sequences from storage and employs tBPTT. As detailed in the Trainer procedure, the long sequence is processed in smaller segments (e.g., 128 steps, corresponding to the model's context length). The final hidden state from one segment is then passed to the next ($h_k \gets h_{k-1}.\text{detach}()$), allowing the model to build a memory that spans thousands of timesteps while keeping gradient computation managable. The complete workflow is detailed in Algorithm \ref{alg:framework}.

Our experimental hardware consisted of a dedicated training node with eight NVIDIA A800 GPUs (one per trainer worker) and three data collection nodes with two NVIDIA 3090 GPUs each (one GPU per rollout worker). We leveraged automatic mixed precision (AMP) to accelerate training. This distributed setup sustained a throughput of approximately 500 environment frames per second (FPS), with each experiment requiring about three days to run.
\newpage

\section{Evaluation Protocols}

\subsection{Minecraft Evaluation}

\paragraph{Benchmark Choices}
To evaluate our model and its baselines, we define three task groups of progressively increasing difficulty: \textit{semantic understanding}, \textit{visible instance interaction}, and \textit{invisible instance interaction}.
For a rigorous evaluation, both our model and the baselines are subjected to the identical conditions specified within each task group.

The first group, semantic understanding, is adapted from the Mine tasks in MCU~\citep{mcu}. These tasks only require the agent to correctly interpret and follow language-based instructions.

The second group, visible instance interaction, is based on the Minecraft Interaction Benchmark~\citep{cai2024rocket}. Here, the agent must not only understand the instruction but also successfully locate and interact with the correct object instance (e.g., ``the sheep on the right'').

The third and most challenging group, invisible instance interaction, is generated by our novel task synthesis pipeline. These tasks introduce several distinct difficulties:
\begin{itemize}
    \item \textbf{Exploration under pressure:} The target instance is often not visible from the agent's spawn point, demanding that the agent explore the environment using visual cues. A tight time limit of 600 steps (approximately 30 seconds) makes efficient exploration critical, as a wrong turn can lead to failure.
    \item \textbf{Complex, game-like scenarios:}  The generated environments are designed to mimic authentic gameplay. Agents must contend with emergent challenges such as switching between tools, handling nearby hostile mobs, and navigating complex terrains and biomes.
    \item \textbf{Challenging skill requirements:} The tasks may require skills, like archery, that pre-trained models often fail to demonstrate, despite the presence of these skills in the training data.
\end{itemize}

\subsection{Unseen Environments Evaluation}

\begin{table}[h]
\setlength{\tabcolsep}{1.0 mm}
\centering
\caption{ \textbf{Bridging the Minecraft Action Space and Other 3D Games.} ``/'' denotes the masked action. } \label{tab:action_mapping} 
\renewcommand{\arraystretch}{1.0}

\footnotesize
\begin{adjustbox}{width=\linewidth}
\begin{tabular}{@{}llll@{}}
\toprule
\textbf{Minecraft} & \textbf{DeepMind Lab} & \textbf{Robot Car} & \textbf{Unreal} \\ \midrule
$\text{forward}=1$ & $a[3]=1$     & 0.1 meters forward   & $\text{velocity}=+100$  \\
$\text{back}=1$    & $a[3]=-1$    & 0.1 meters backward  & / \\
$\text{left}=1$    & $a[2]=-1$    & 0.1 meters left & / \\
$\text{right}=1$    & $a[2]=1$    & 0.1 meters right  & $\text{velocity}=-100$ \\
$\text{use}=1$    & /    & /  &  /\\
$\text{attack}=1$  & $a[4]=1$     & trigger beeper   & $\text{pick}=1$  \\
$\text{yaw}=x$     & $a[0]=4.75x$ & $\text{yaw by wheels}=x$ & $\text{angular}=x$     \\
$\text{pitch}=x$   & $a[1]=2.78x$ & $\text{pitch for camera}=x$         & $\text{viewport}=x$             \\ 
$\text{jump}=1$   & / & /      & /          \\ 
$\text{sneak}=1$   & $a[6]=1$ & /          & /            \\ 
$\text{composite actions}$   & the same time & sequential          & the same time            \\ 

\bottomrule
\end{tabular}
\end{adjustbox}
\end{table}

\paragraph{Action space mapping}

We facilitate the agent's application in novel environments by constructing a rule-based action mapping (Table \ref{tab:action_mapping}). Critically, this method \textbf{obviates the need for environment-specific fine-tuning}, as our trials demonstrated that such this approach is quite insensitive to the choice of action mapping.

\paragraph{Unreal Zoo Rescue Task}
For this task, we adapt the Level 3 environment from the ATEC Challenge in Unreal~\citep{unrealzoo}. In this scenario, the agent must identify injured individuals by interpreting surrounding visual cues, pick them up, and transport them to designated stretchers—a process that demands strong spatial reasoning abilities. 
Images of the injured person serve as prompts for our model.
Furthermore, this Unreal Engine environment provides observations at a $640\times480 $ resolution, a notable deviation from the 640x360 resolution of the Minecraft training data. This discrepancy serves as a key test of the agent's robustness and its ability to generalize across different visual domains.
The agent is rewarded in two stages: 0.5 for retrieving an injured person and 0.5 for the successful transfer.

\paragraph{DMLab30 Fruit Collection}
This task is set in the \texttt{explore\_object\_locations\_small} environment from DMLab30~\citep{dmlab}. The agent must collect fruits within 300 steps, following human-generated prompts curated from live gameplay.

\begin{figure}[ht]
\begin{center}
\includegraphics[width=0.99\linewidth]{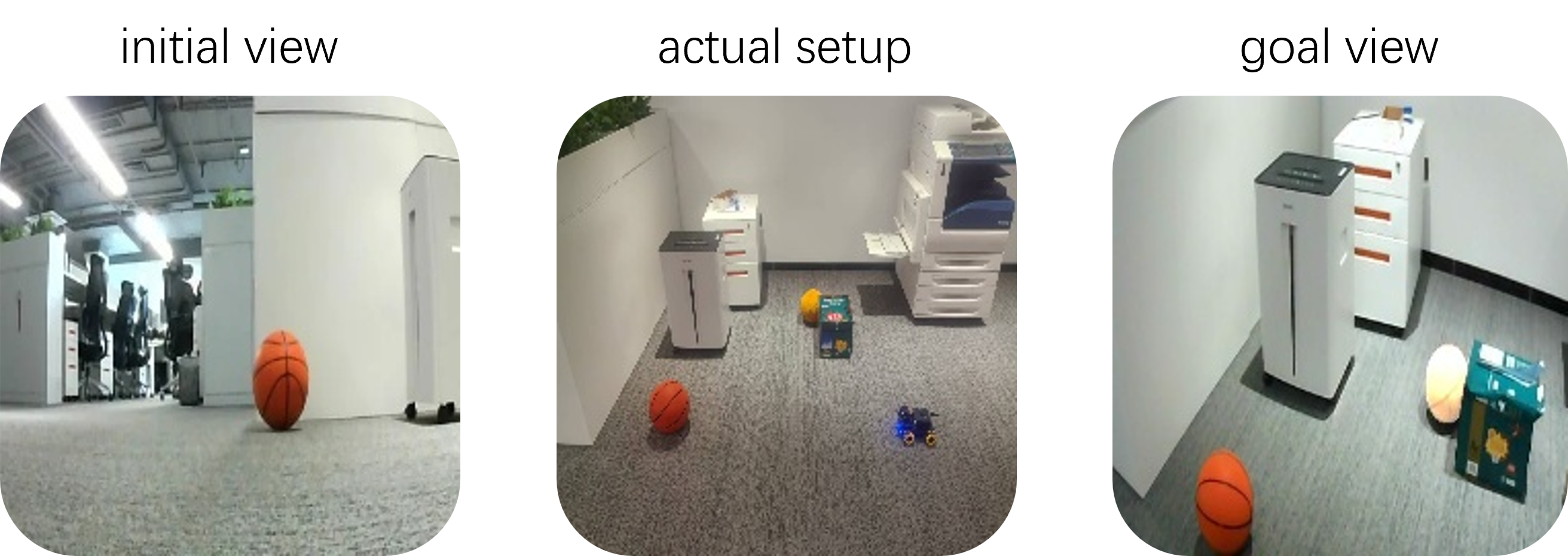}
\end{center}
\caption{
\textbf{The zero shot setting for real world environments.} The goal would be blocked by the paper box if the car naively rotates towards the direction.
}
\label{fig:car}
\end{figure}

\begin{figure}[ht]
\begin{center}
\includegraphics[width=0.99\linewidth]{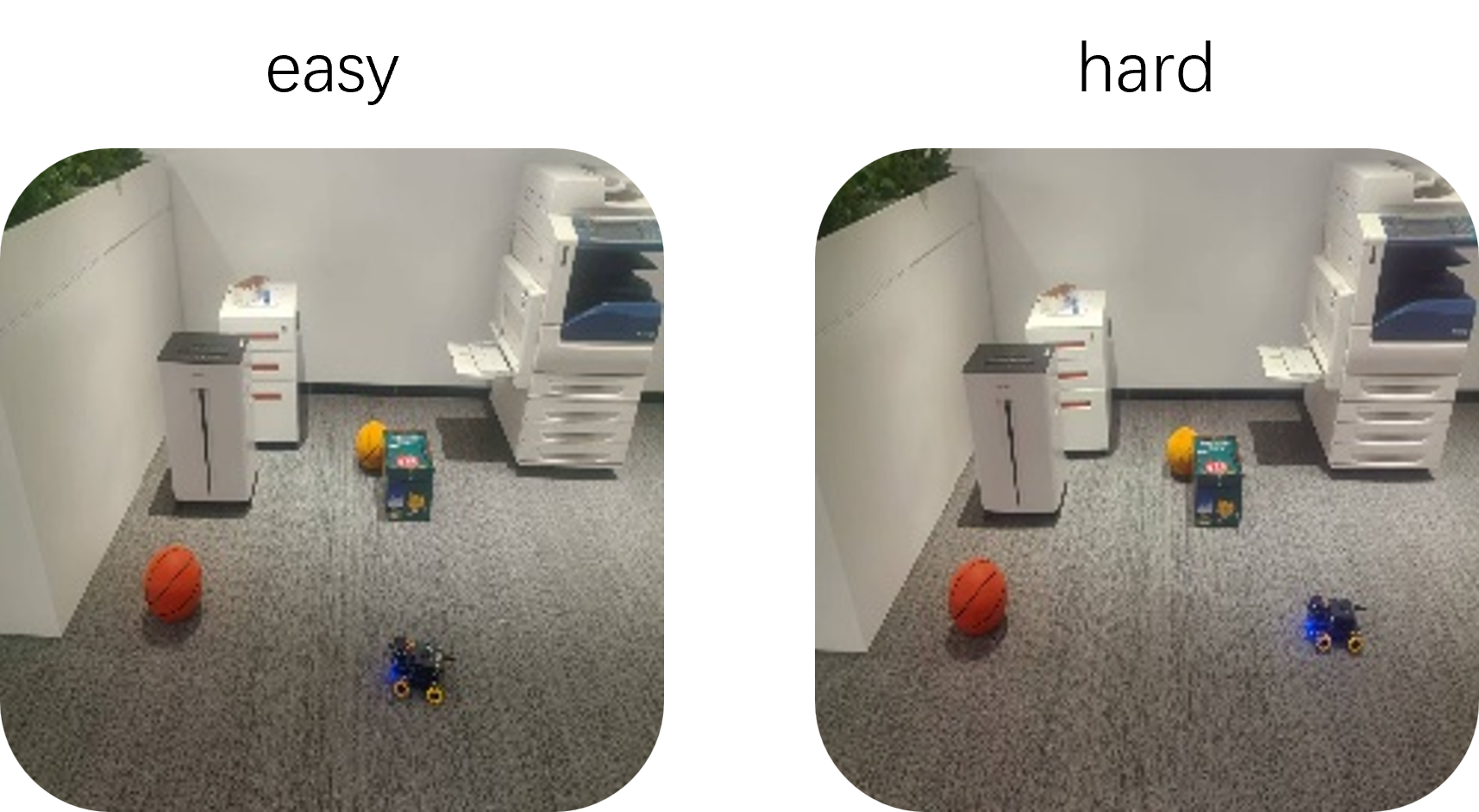}
\end{center}
\caption{
\textbf{The easy and hard variant of cross-view approach setting.
}
}
\label{fig:level_comparison}
\end{figure}

\begin{figure}[ht]
\begin{center}
\includegraphics[width=0.99\linewidth]{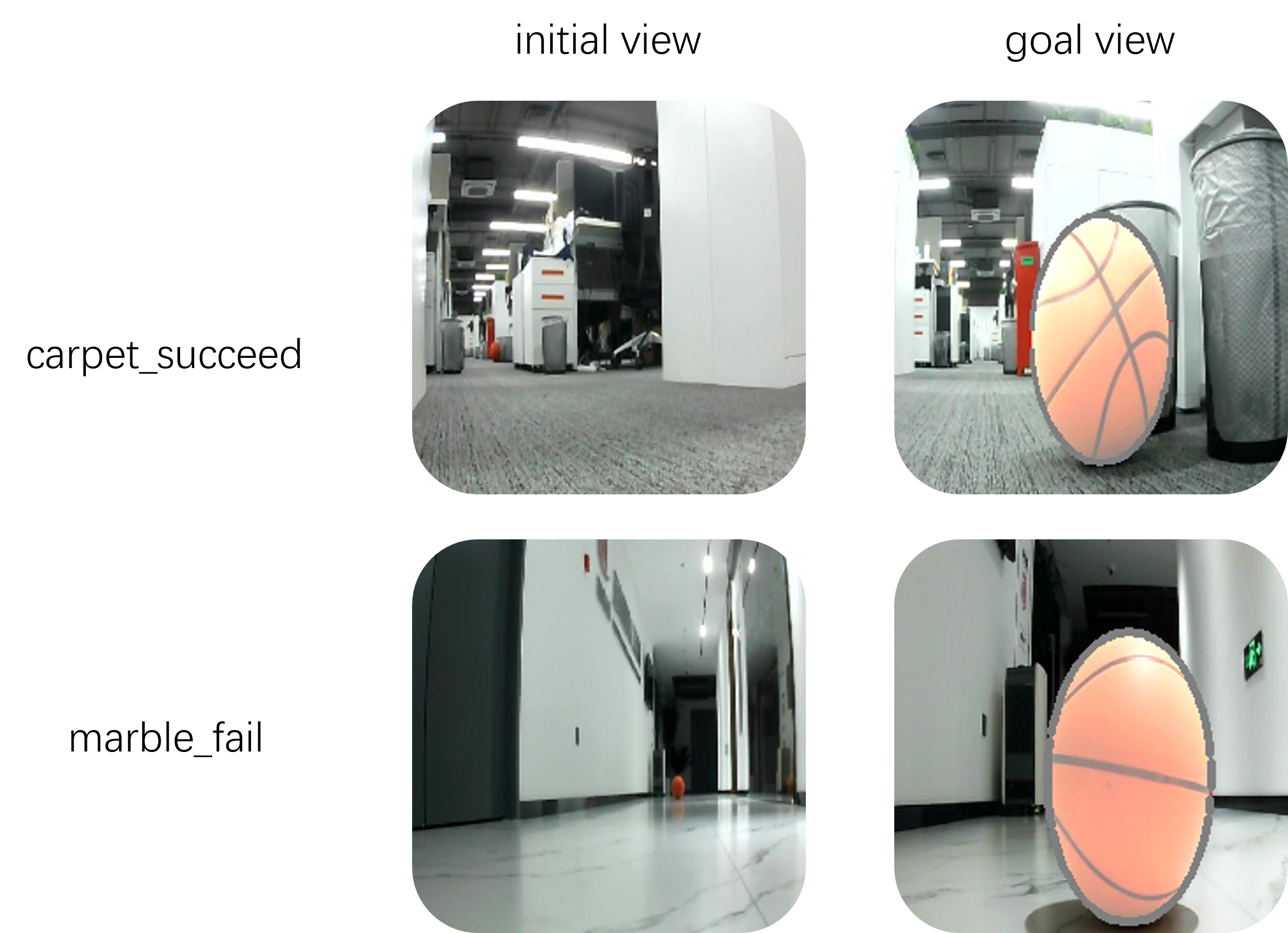}
\end{center}
\caption{
\textbf{A long distance approach task.
} The agent fails in the marble hallway due to Out Of Distribution challenges and perform better in the indoor case.
}
\label{fig:other_settings}
\end{figure}

\begin{figure}[ht]
\begin{center}
\includegraphics[width=0.99\linewidth]{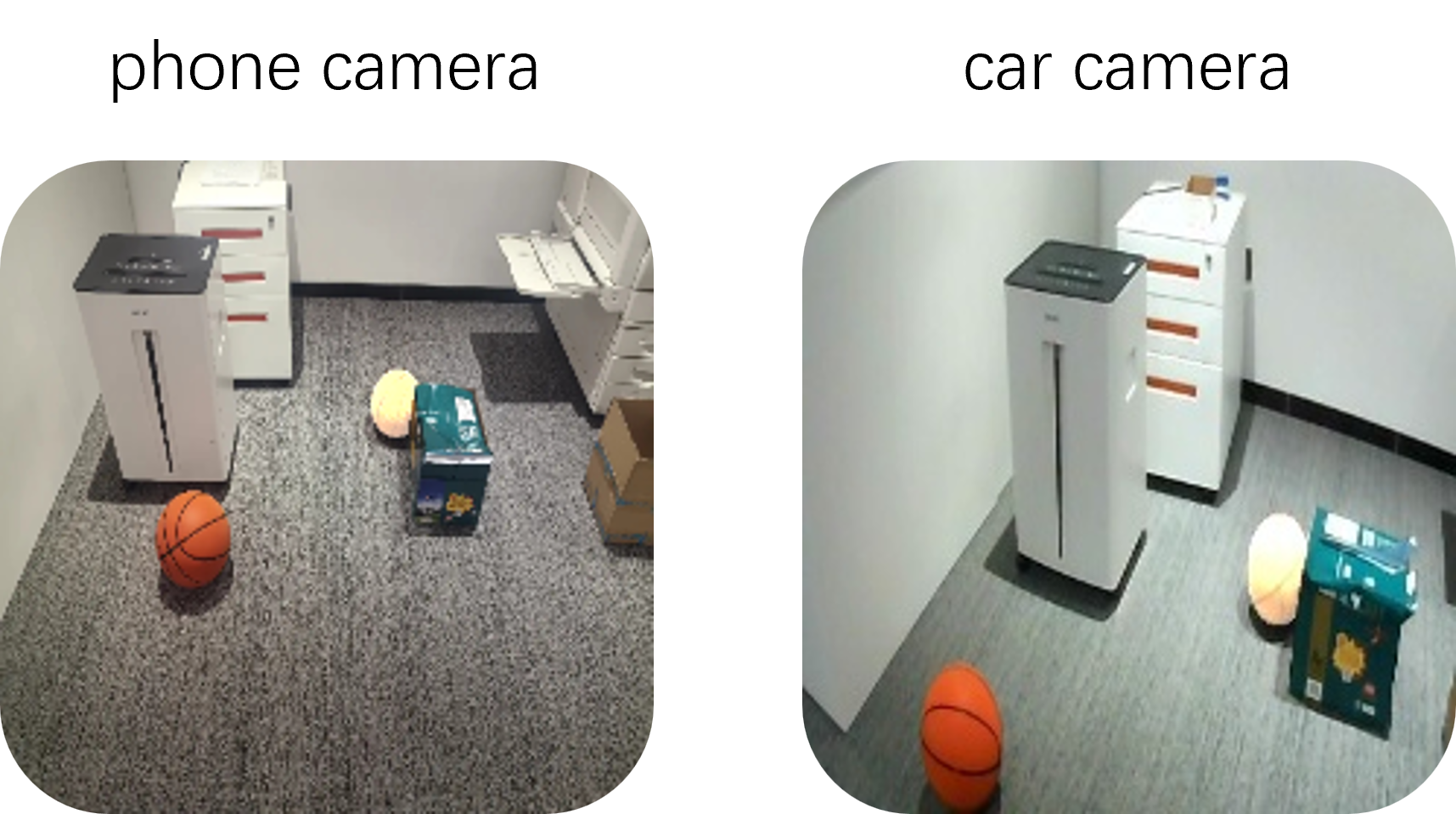}
\end{center}
\caption{
\textbf{Different goal captures.
} Goals from phone cameras does not deteriorate the performance of our method.
}
\label{fig:other_goals}
\end{figure}
\subsubsection{Real World Experiments} 

\paragraph{Environment Protocols}Our real world experiments are conducted indoors, using a remote inference server (one NVIDIA 4090 GPU) that synchronously transmits computed actions to the robot car. The car remains blocked while awaiting results. Once received—whether a single command or a set of actions (e.g., yaw, pitch, forward)—the actions are executed sequentially. After completion, an onboard camera image of $640\times360 $ resolution is sent back to the server for the next inference step. To align with the motion control scheme used in Minecraft, the vehicle’s wheel motors control both translational and yaw movements, while a dedicated camera motor adjusts pitch. \textbf{We did not intentionally choose the forward distance.} Due to the latency in mechanical execution and stabilization, the vehicle operates at a control frequency of 2Hz, significantly slower than Minecraft’s 15+Hz frame rate.

\paragraph{Primary Cross-View Goal}  
We propose a \textbf{cross-view approach setting} shown in Figure~\ref{fig:car}, including the initial image observed by the agent, global environment layout, and the goal image. The goal image is captured from a top-down perspective by holding the car in the air. In the easy variant shown in the left of Figure~\ref{fig:level_comparison}, a simple rightward yaw suffices to bring the yellow ball into view; both ROCKET-2 and our method succeed reliably, with exhibiting a slightly higher short-range success rate. In the hard variant, the ball is occluded by a paper box, forcing the car to detour around the obstacle and then reorient its viewpoint. ROCKET-2 frequently stalls: rotating in place without progress and succeeds in only 3 of 12 trials. In contrast, our method shows clear recovery behavior and active re-planning: it completes the detour from both the left and right sides in 8 of 12 trials. Three trajectories begin with substantial deviations (e.g., navigating outside the goal frame), but subsequently realign toward the target and succeed, demonstrating that early errors do not preclude eventual task completion. We masked the pitch action for simplicity and observed negligible difference in performance.

\paragraph{Additional Variants}  We also evaluated several other settings, including an alternative goal image(Figure~\ref{fig:other_goals}) and a long distance approach task(Figure~\ref{fig:other_settings}) in different environment layouts. The goal captured from the phone with different lighting and camera parameters does not deteriorate the performance. Further breakdowns of successes and failure modes under these conditions are provided in the failure analysis section. 

\section{Model Architecture}

We use ROCKET-2~\citep{cai2025rocket} as the pre-trained model.
ROCKET-2 is designed to align goals across different views. It processes training trajectories, each containing a global condition ($c_g$), a sequence of visual observations ($o_t$), and their corresponding segmentation masks ($m_t$) over time $t$. A specific time step $g$ with a valid mask is selected as the cross-view reference. For consistency, all visual inputs ($o_t$) and masks ($m_t$) are resized to $224\times224$ pixels.

First, ROCKET-2 extracts features from the visual data:
Each visual observation $o_t$ is processed by a frozen DINO-pretrained ViT-B/16~\citep{vit,dino} (Vision Transformer, Base architecture, 16x16 pixel patches). This encoder outputs 196 visual tokens, denoted as $\{\hat{o}_t^{i}\}_{i=1}^{196}$. For computational efficiency, this ViT-B/16 encoder remains frozen during the entire training process.
Separately, each segmentation mask $m_t$ is encoded using a trainable ViT-tiny/16, which also produces 196 tokens, $\{\hat{m}_t^{i}\}_{i=1}^{196}$.

Next, the model integrates information from the cross-view reference ($o_g, m_g$) to ensure spatial alignment.
It combines the encoded visual tokens and mask tokens by concatenating their channels, and then processes them by a Feed-Forward Network (FFN) to create a fused spatial representation $h_g^{i}$.

$h_g^{i}$ is then used in a non-causal Transformer encoder, which takes the current visual tokens and this fused cross-view condition as input.
By concatenating these into a sequence of 392 tokens, this `SpatialFusion' step combines spatial details from the current view with the cross-view reference, producing a detailed frame-level representation $x_t$.

After obtaining the frame-level representation $x_t$, a causal TransformerXL architecture captures temporal relationships between frames, resulting in a rich temporal representation $f_t$.
Finally, $f_t$ is fed into a lightweight network responsible to predict the action ($\hat{a}_t$), centroid ($\hat{p}_t$), and visibility ($\hat{v}_t$) at the current time step. The model's training is guided by a negative log-likelihood loss function, which is summed over all time steps for each episode $n$, effectively acting as a cross-entropy-like loss to minimize the discrepancy between predicted and ground truth values:
\begin{equation*}
\mathcal{L}(n) = \sum_{t=1}^{L(n)} -a_t^n \log \hat{a}_t^n - p_t^n \log \hat{p}_t^n - v_t^n \log \hat{v}_t^n.
\end{equation*}

\section{Analyzing Failure Cases}

We conduct a detailed analysis of failure cases in both Minecraft and unseen environments.

\paragraph{Minecraft} Three primary reasons lead to these failures:
\begin{itemize}
\item \textbf{Occasional Segmentation Issues:} This issue stems from several factors, including the fact that SAM (Segment Anything Model) is not specifically trained for Minecraft environments, and the presence of occlusions from elements like the message bar or the agent's hands, which obstruct objects. However, as vision-language models continue to improve, these challenges are expected to be effectively resolved.

\item \textbf{Insufficient Visual Cues:} Certain cross-view scenarios fail to provide adequate visual cues necessary for task completion. This necessitates extensive exploration, leading to high failure rates within limited timeframes.

\item \textbf{Lack of Incentive for Latent Skills:} Although certain latent skills—such as pillar jumping, shield defense against hostile mobs, or parkouring—may exist in the pre-trained models, they are not incentivized or reinforced during the RL process. Consequently, these abilities remain latent and are rarely exhibited by the agents when required.
\end{itemize}

\paragraph{Unreal Zoo Rescue Task}
The failure in the Unreal Zoo Rescue Task can be attributed to several factors. First, in highly complex environments, agents often struggle with accurate spatial reasoning, making it difficult to navigate and complete objectives. Second, certain necessary skills—such as opening doors—are not present in Minecraft and thus are absent from the agent’s repertoire; addressing these gaps may require test-time training or fine-tuning. Finally, issues such as unintended collisions or “clipping” through objects also contribute to unsuccessful task completion.

\paragraph{DMLab30 Fruit Collections} The failure in the DMLab30 Fruit Collections task stems from several key issues. First, the low distinctiveness of DMLab30’s environments makes it difficult for the agent to distinguish between different observations, leading to confusion during navigation. Additionally, agents sometimes get stuck in dead ends, likely due to discrepancies between the environment dynamics of DMLab30 and those of Minecraft. Interestingly, for the pre-trained ROCKET-2 agent, the primary cause of failure is its difficulty in accurately shooting the fruit, suggesting that ROCKET-2 lacks robustness to subtle skill differences, which hinders effective transfer.

\paragraph{Real World Experiments} 

Our method suffers from severe OOD challenges in the real world. First, the discrepancy in camera viewpoints. In Minecraft, the agent perceives the world from an elevated, human-like perspective, whereas in the real-world robotic platform, the onboard camera is mounted at a much lower height. This results in severe perspective distortions and fundamentally different visual distributions. For example, a large portion of the frame is often occupied by the floor or monotonous white walls, which affects depth perception and spatial reasoning. Second, the dynamics of the real world are subtly different from Minecraft especially near objects, such as the forward distance, collision and higher chances to get stuck when turning. These two factors deteriorates the model performance. 

When translated back into Minecraft units, the cross-view setting corresponds to \textbf{easy} difficulty level: optimal control sequences require only about 30 steps to reach the goal. However, our real-world policies exhibit lower success rates, reduced stability, and less efficient trajectories compared to their Minecraft counterparts. Though the recovery capability of our method differentiates it from ROCKET-2, \textbf{suboptimal exploratory behaviors occur more frequently}, suggesting a higher likelihood of deviation from the shortest or most direct paths.

Other failure cases in longer approach tasks stem from the observation mismatch. The longer approach task (Figure~\ref{fig:other_settings}) places the goal basketball directly in front of the car, but with more than 10 meters away; the goal image is shot 0.5 meters before the ball, while the goal is only 30 pixels wide in the initial $640\times480$ observation. Despite the absence of obstacles, the agent repeatedly engaged in inefficient behaviors, alternating between brief forward movements and 360 degree spinning in place. This was particularly evident in the hallway, with white marble floors, white walls and bright lighting, which we hypothesize caused the image input to fall too far outside the distribution encountered during training. Notably, When the same experiment is conducted in an office corridor with textured gray carpet, the car’s exploration remains more focused and directed even with more sideways. \textbf{In all long approach tasks however, the agent rarely takes straight trajectories.} These findings also revealed that our current model, although effective in short-range navigation and fine-grained corrections, performs poorly in sparse and visually homogeneous settings such as long hallways, and the reaction-based policy does not guarantee efficiency. We hypothesize that explicit spatial planning might relieve these issues.

\section{Demonstration Showcases}

\begin{figure*}[htbp]
    \centering
    \begin{subfigure}{\linewidth}
        \centering
        \includegraphics[width=0.85\linewidth]{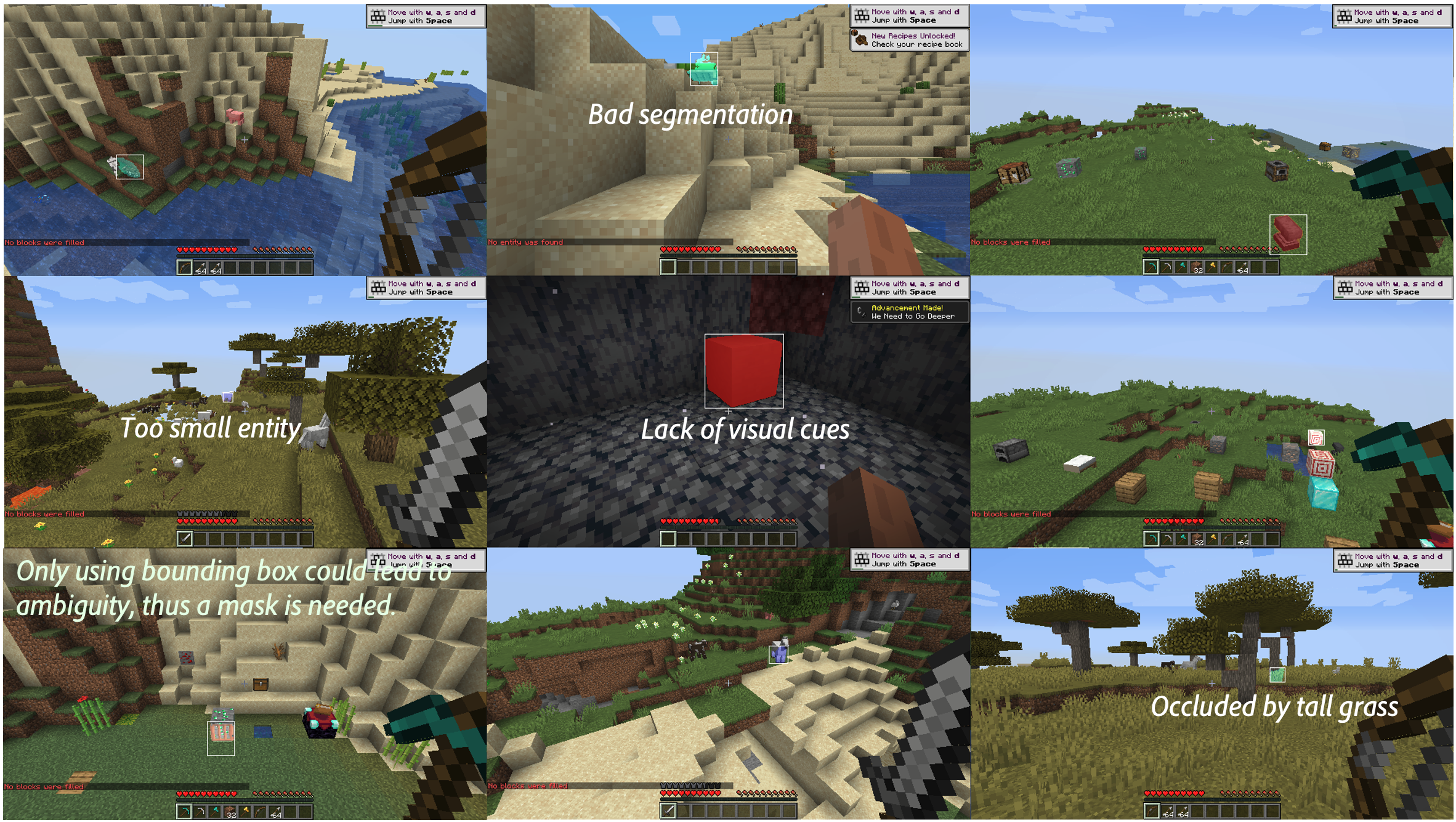}
        \caption{\textbf{Gallery of task synthesis results}, illustrating both successful cases and challenging corner cases.}
        \label{fig:goal_gallary}
    \end{subfigure}
    \vspace{1em}
    \begin{subfigure}{\linewidth}
        \centering
        \includegraphics[width=0.85\linewidth]{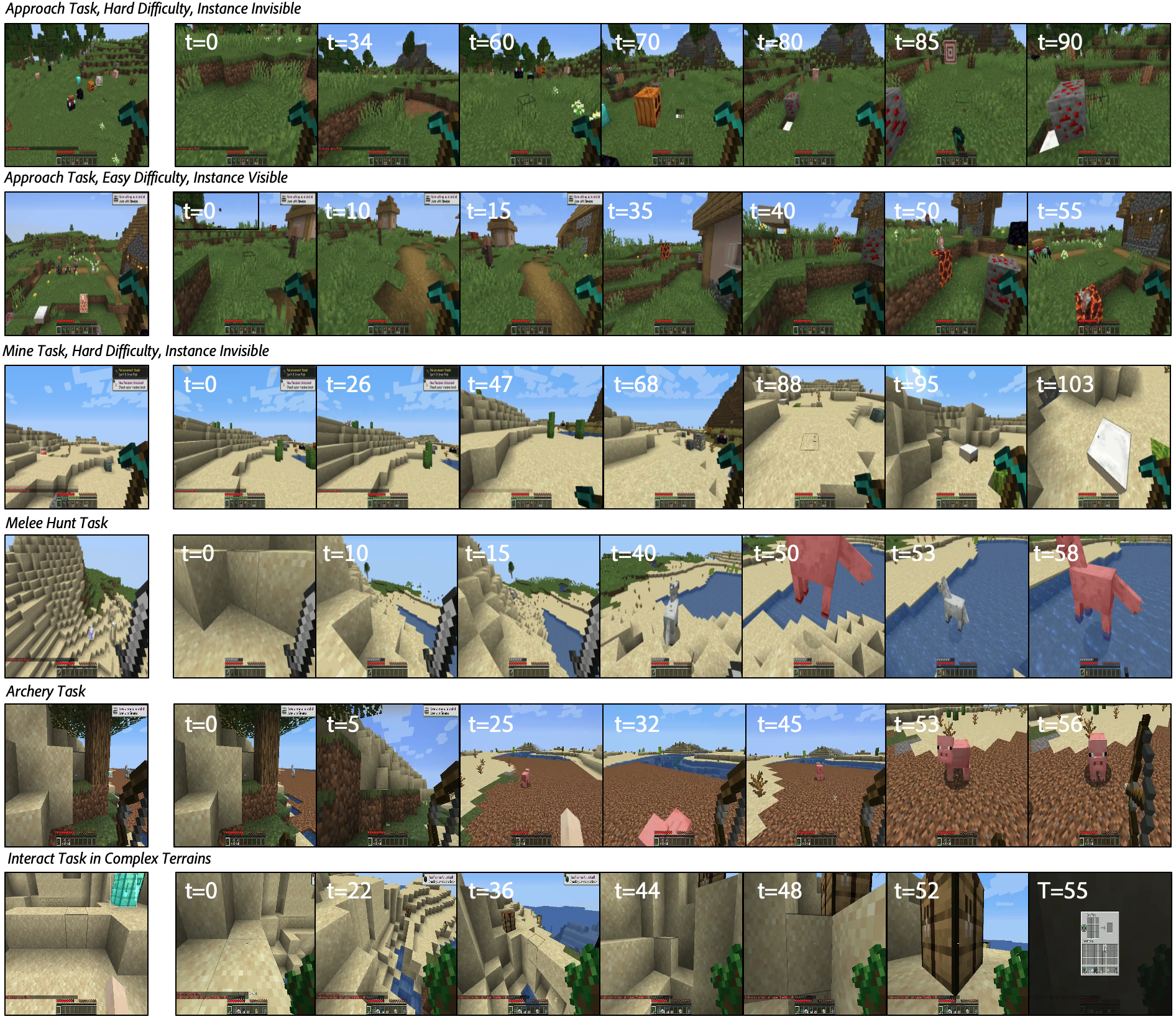}
        \caption{\textbf{Minecraft Demonstrations.} We present demonstrations on the \textit{Approach}, \textit{Break}, \textit{Interact}, \textit{Melee Hunt}, and \textit{Archery} tasks. Additionally, we compare performance on \textit{Hard} versus \textit{Easy} tasks, noting that \textit{Hard} tasks typically require exploration guided by visual cues.}
        \label{fig:minecraft_demos}
    \end{subfigure}
        \caption{(a) Gallery of task synthesis results. (b) Minecraft Demonstrations.}
    \label{fig:combined}
\end{figure*}

\begin{figure*}
    \centering
    \includegraphics[width=\linewidth]{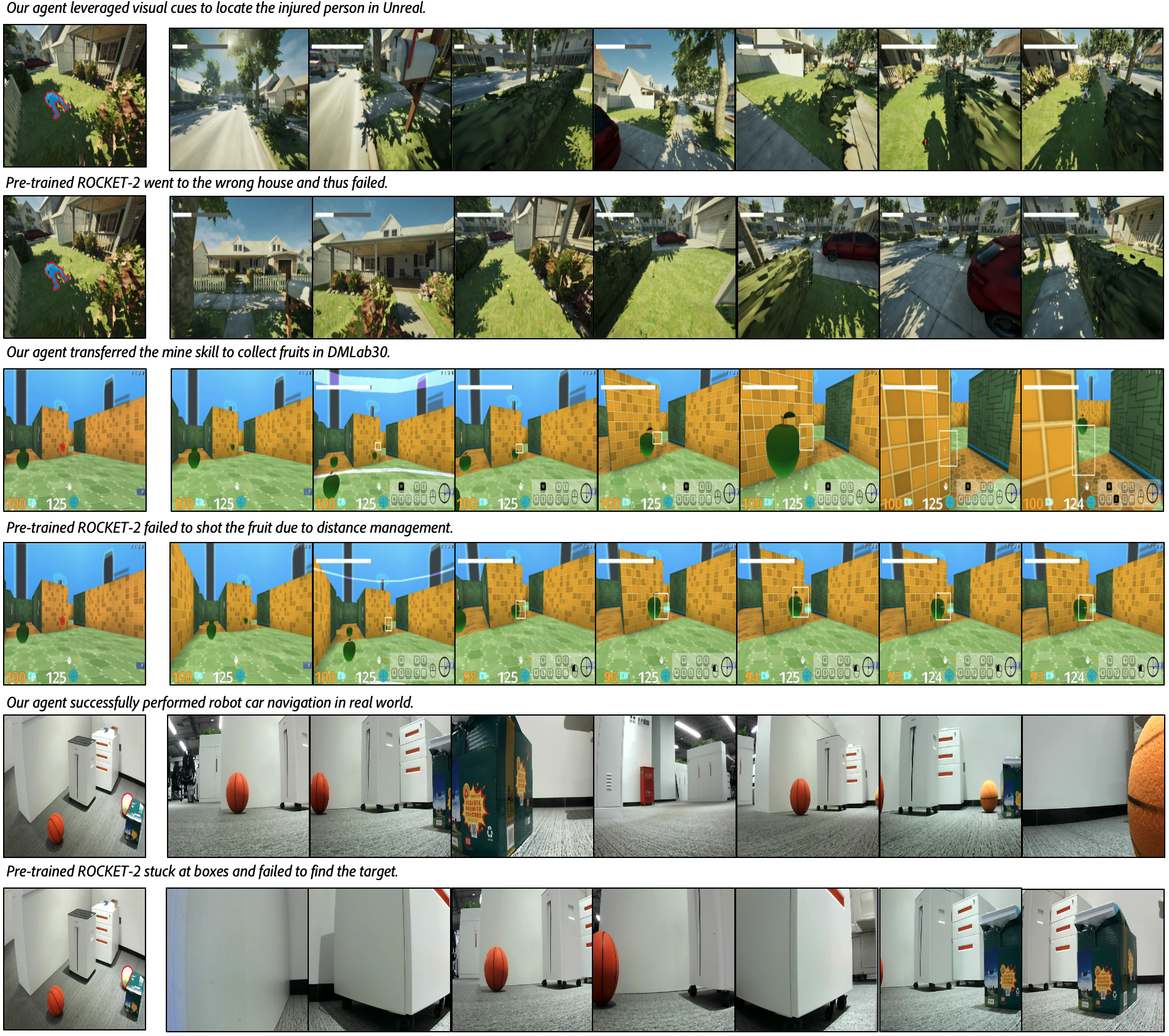}
    \caption{\textbf{Zero-Shot Environment Showcases.} We evaluated both the pre-trained ROCKET-2 and our agent in Unreal~\citep{unrealzoo}, DMLab30~\citep{dmlab}, and real-world environments. Experimental results demonstrate that this reinforcement learning approach can significantly improve performance even in unseen settings.}
    \label{fig:zero_shot_envs}
\end{figure*}

We provide visualizations of our demonstrations in Figure \ref{fig:minecraft_demos} and Figure \ref{fig:zero_shot_envs}.
Comparisons between \textit{Hard} and \textit{Easy} tasks are also illustrated in Figure~\ref{fig:minecraft_demos}, where, in \textit{Hard} tasks, relevant instances are often not directly visible.
We also present a gallery showcasing both successful examples and challenging corner cases from our task synthesis results. Occasional issues such as poor segmentation or insufficient visual cues can increase the difficulty of training and evaluation. However, as vision-language models continue to advance, we expect these challenges to be effectively addressed.
Some cases further highlight the necessity of using SAM, as relying solely on bounding boxes for masks can result in occlusion and ambiguity.

\end{document}